\pdfoutput=1

\documentclass[11pt]{article}

\usepackage[table]{xcolor}
\usepackage{tcolorbox}
\usepackage{amsmath}
\usepackage{multirow}
\usepackage{setspace}
\usepackage{booktabs}

\usepackage[preprint]{acl}

\usepackage{times}
\usepackage{latexsym}

\usepackage[T1]{fontenc}

\usepackage[utf8]{inputenc}

\usepackage{microtype}

\usepackage{inconsolata}

\usepackage{graphicx}

\title{Investigating Context Faithfulness in Large Language Models: The Roles of Memory Strength and Evidence Style}

  \author{Yuepei Li,\hspace{2mm} Kang Zhou,\hspace{2mm} Qiao Qiao,\hspace{2mm} Bach Nguyen,\hspace{2mm} Qing Wang,\hspace{2mm} Qi Li \\
  Department of Computer Science \\
  Iowa State University, Ames, Iowa, USA \\
  \texttt{\{liyp0095, kangzhou, qqiao1, ntbach, qingwang, qli\}@iastate.edu} \\}

\begin{document}
\maketitle
\begin{abstract}

Retrieval-augmented generation (RAG) improves Large Language Models (LLMs) by incorporating external information into the response generation process. However, how context-faithful LLMs are and what factors influence LLMs' context faithfulness remain largely unexplored. In this study, we investigate the impact of memory strength and evidence presentation on LLMs' receptiveness to external evidence. We quantify the memory strength of LLMs by measuring the divergence in LLMs' responses to different paraphrases of the same question, which is not considered by previous works. We also generate evidence in various styles to examine LLMs' behavior. Our results show that for questions with high memory strength, LLMs are more likely to rely on internal memory. Furthermore, presenting paraphrased evidence significantly increases LLMs' receptiveness compared to simple repetition or adding details. These findings provide key insights for improving retrieval-augmented generation and context-aware LLMs. Our code is available at \href{https://github.com/liyp0095/ContextFaithful}{https://github.com/liyp0095/ContextFaithful}.

\end{abstract}

\section{Introduction}





Retrieval-Augmented Generation (RAG) \cite{fan2024survey, zhao2023retrieving} has gained increasing popularity as it improves the performance of Large Language Models (LLMs) by integrating external information during the generation process, particularly when the model’s internal knowledge is insufficient or outdated \cite{bianchini2024enhancing, procko2024graph, siriwardhana-etal-2023-improving, 10561020, wang2024biorag, jeong2023generative}. It raises the importance of the study of how context-faithful LLMs are. 
In this study, we explore whether LLMs are context-faithful when encountering external information, particularly when that information conflicts with the LLMs' internal memory.

\begin{figure}[t]
    \centering
    \includegraphics[width=\linewidth]{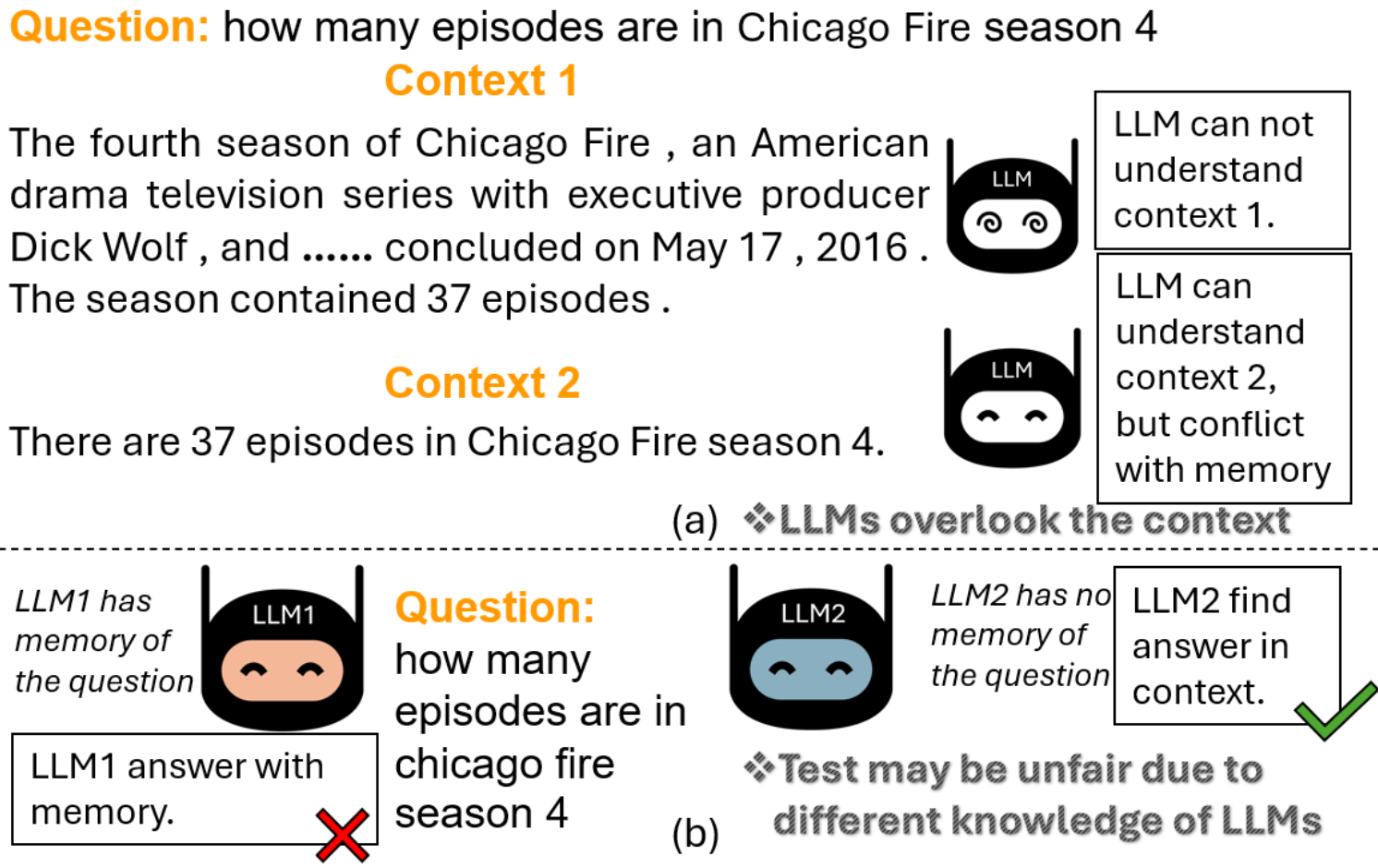}
    \caption{Demonstration of issues in context faithfulness testing schema: a) LLMs may fail to comprehend long contexts, resulting in low receptiveness to the context. b) Evaluating LLMs on the same dataset may be unfair due to variations in their knowledge.}
    \label{fig:motivation}
\end{figure}

To investigate the issue of context faithfulness, there are two main approaches to creating knowledge conflict contexts. One approach \cite{longpre2021entity, chen-etal-2022-rich} is entity substitution, which replaces the gold entity in context with a similar one. 
Another approach \cite{xie2024knowledgeconflict, jin-etal-2024-tug} involves generating counter-memory evidence with LLMs, and these studies have shown that LLMs are generally receptive to external evidence as long as it is coherent.

These methods, however, overlook some important aspects of the task. First, previous work \cite{longpre2021entity, xie2024knowledgeconflict} provides long contexts to LLMs, which can be challenging for LLMs to understand \cite{xie2024knowledgeconflict}. It makes the test difficult to distinguish whether the cause of LLM behavior is due to knowledge conflicts or lacking comprehension ability. For example, in Figure \ref{fig:motivation} (a), the LLM may overlook both contexts, but the reason why Context 1 is ignored could be a lack of comprehension rather than knowledge conflicts. 
Second, different LLMs are trained with different data and are likely to obtain different knowledge. Thus, testing LLMs on the same dataset may be unfair. The test may favor LLMs with less knowledge. As shown in Figure \ref{fig:motivation} (b), 
LLMs with strong memory are less likely to be correct. 


To address these issues, we introduce a method to quantify the memory strength of LLMs and generate short evidence in various styles to examine LLMs' behavior. 
Inspired by \citet{zhao-etal-2024-knowing}, we assess memory strength by measuring the divergence in LLMs' responses to different paraphrases of the same question. Intuitively, an LLM demonstrates high memory strength when it consistently provides the same answer across all paraphrased versions of a question. For evidence styles, we classify them into direct and indirect forms: direct evidence provides a straightforward answer to the question, while indirect evidence incorporates additional details to support the answer. 
Unlike \citet{wan-etal-2024-evidence} that focuses on how different types of evidence influence LLMs' behavior, we study how the presentation of evidence affects its convincing availability. 
Through these methods, we analyze the relationship between context faithfulness and LLM memory strength, and we explore the impact of different evidence styles on context faithfulness. Our conclusions are as follows:
\begin{itemize}
    \item The receptiveness of LLMs to external evidence is strongly correlated with memory strength to the question. We observed this relationship both across different datasets and different LLMs. Contrary to the findings of \cite{xie2024knowledgeconflict, jin-etal-2024-tug} that LLMs are highly receptive (less than 5\%) to external evidence when it is coherent, we find that the probability of the model relying on its internal memory is non-negligible for questions that the LLMs have a strong memory. For example, GPT-4, which has strong memory on the NQ dataset, answers almost 50\% of the questions with internal memory. We also demonstrate an urgent need for memory strength-aware evaluation metrics.
    \item The style of the evidence plays an important role in LLM’s receptiveness to external information. Our research demonstrates that presenting the LLM with multiple paraphrases of the same evidence substantially increases its receptiveness. This approach outperforms simple repetition of the evidence and is more effective than adding additional details to the evidence. These findings provide valuable insights to the research of RAG.
\end{itemize}

\section{Related Work}

\begin{table*}[t]
\centering
\resizebox{0.95\textwidth}{!}{
\begin{tabular}{|p{3.0cm}|p{8.0cm}|p{8.0cm}|}
\hline
\textbf{Origin Question} & \textbf{Paraphrased Questions} & \textbf{Answers with LLaMA2-7B} \\ \hline
\multirow{4}{3.0cm}{Who sings the Christmas song ``Mary Did You Know''?} 
&  Can you tell me the artist who performs the Christmas song ``Mary Did You Know''? & Of course! The artist who performs the Christmas song ``Mary Did You Know'' is \colorbox{blue!30}{Mark Lowry}. \\ \cline{2-3} 
&  Who is the singer behind the Christmas song ``Mary Did You Know''?              &  The singer behind the Christmas song ``Mary Did You Know'' is \colorbox{blue!30}{Mark Lowry}. \\ \cline{2-3} 
&  Which musician sings the Christmas tune ``Mary Did You Know''?                  &  \colorbox{red!30}{Brandy Clark} \\ \cline{2-3} 
&  Who is the vocalist that sings ``Mary Did You Know'' during the holiday season? &  The vocalist who sings ``Mary Did You Know'' during the holiday season is \colorbox{blue!30}{Mark Lowry}. \\ \hline
\end{tabular}
}
\caption{LLMs give different answers when asked with paraphrased questions.}
\label{tab:qa}
\end{table*}

\subsection{Context Faithfulness of LLM}

To update static factual knowledge \cite{lazaridou2021mind, karpukhin-etal-2020-dense, NEURIPS2023_9941624e} in LLMs, the retrieval-based method has been introduced to involve external information to LLMs \cite{lazaridou2022internet, 10.5555/3648699.3648950, khattab2022demonstrate, santhanam-etal-2022-colbertv2, gao-callan-2022-unsupervised}. However, these methods can introduce \textbf{knowledge conflicts} between the introduced external information (context) and pre-existing internal memories from LLMs. LLMs often rely on their internal memories, and overlook the contextual evidence \cite{longpre2021entity}. To make LLMs more faithful to context, recent studies \cite{neeman-etal-2023-disentqa, li-etal-2023-large} fine-tune LLMs on counterfactual contexts, where the original facts are replaced with counterfactual ones. Another work \cite{zhou-etal-2023-context} proposes a novel approach using prompting to improve context faithfulness in LLMs without additional fine-tuning. 
Recent works \cite{farahani2024deciphering, wadhwa2024rags} found that models like T5, LLaMA2-7B, and Phi2-2.7B are generally context-faithful. However, they did not evaluate newer models (e.g., LLaMA3.2, GPT-4) or consider the impact of memory strength, limiting the generality of their conclusions.

A related area of research focuses on \textbf{prediction with abstention}. \citet{neeman-etal-2023-disentqa, zhou-etal-2024-gendecider} introduces answerability augmentation, where LLMs are trained to respond with "Unanswerable" when presented with irrelevant or randomly generated contexts. This ensures that the models do not make incorrect predictions without reliable evidence. Further studies \cite{wang-etal-2023-extracting, wang-etal-2022-rely} develop confidence calibration techniques to improve context faithfulness by encouraging LLMs to avoid overly confident predictions in ambiguous or uncertain situations.

In our work, we investigate the context faithfulness of LLMs when faced with conflicting knowledge. We define a model as context-faithful if it demonstrates high receptiveness to new facts and evidence with strong conflicting memories. This capability is essential to ensure the reliability of LLMs in the RAG system. 

\subsection{Construction of Knowledge Conflicts}

In controlled experiments, knowledge conflicts are typically simulated by constructing counterfactual memories based on a model's internal memory. Various heuristic approaches have been proposed for this purpose, such as negation injection \cite{kassner-etal-2021-beliefbank, petroni2020how, pan2021contraqa} and entity substitution \cite{longpre2021entity, chen-etal-2022-rich, si2023prompting, zhou-etal-2023-context}. Negation injection alters facts by introducing negations and entity substitution replaces mentions or entities in the evidence with alternatives to generate \textbf{counter-fact} evidence. However, these techniques are constrained to word-level edits, which can lead to low coherence across the constructed evidence. To address this limitation, recent studies \cite{xie2024knowledgeconflict, jin-etal-2024-tug} have explored generating evidence using LLMs, producing more coherent and consistent counterfactual content. We adopt this approach in generating our dataset, ensuring the generated evidence maintains a higher level of coherence.

\section{Methodology}







\begin{figure*}[t]
    \centering
    \includegraphics[width=\linewidth]{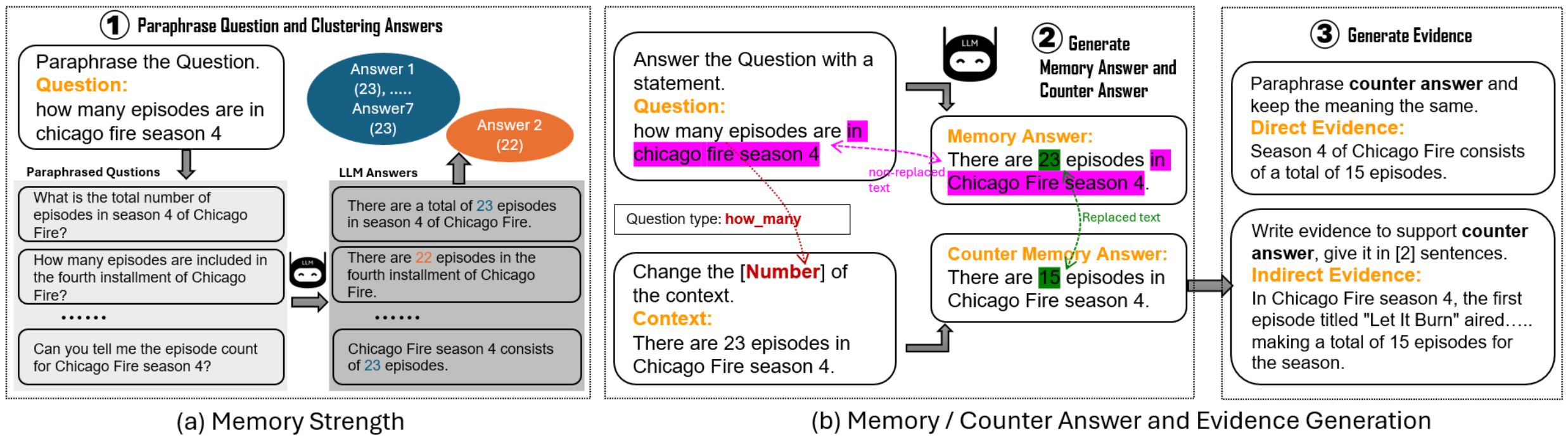}
    \caption{Framework for Evaluating LLMs' Context Faithfulness. 
In Step 1, we calculate the memory strength of a question using the consistency of answers to different paraphrases of the same question. In Step 2, we generate MA (memory answer) under the closed-book setting and CMA (counter memory answer) by modifying \colorbox{teal!70}{\small answer entity} in MA while keeping the \colorbox{violet!70}{\small other information}. In Step 3, we generate supporting evidence for the CMA in various styles. In Step 4 (not shown), we test the LLM's response by presenting the questions with the evidence. All experiments are implemented under the zero-shot setting to avoid the bias introduced by demonstrations.}
    \label{fig:framework}
\end{figure*}

\subsection{Problem Definition}
Following prior work \cite{longpre2021entity, chen-etal-2022-rich, xie2024knowledgeconflict}, we adopt question answering (QA)
task as the testbed for knowledge conflict experiments.
For a given \textbf{question $Q$}, if the answer generated by the LLM relies solely on its internal parameters, it is referred to as the \textbf{memory answer} (MA). If an evidence passage $E$ is provided with question $Q$, then ideally, LLM should generate an answer based on $E$, even if $E$ conflicts with memory answers. We call the answers that conflict with MA as \textbf{counter memory answer} (CMA).


\subsection{Datasets}

We use two datasets for our experiments: the long-tail, entity-based QA dataset popQA, and the popular, human-written question dataset Natural Questions (NQ). Specifically:

\begin{itemize}
    \item \textbf{popQA} \cite{mallen-etal-2023-trust} is an entity-centric question-answering dataset comprising 14,000 questions. The dataset is derived from knowledge triples in Wikidata, where questions are generated using question templates specific to different relationship types. popQA aims to capture a realistic, long-tail distribution of entity popularity, making it a valuable resource for studying the performance of lesser-known entities. \citet{xie2024knowledgeconflict} use popQA to test the receptiveness of LLMs by eliciting high-quality internal memory from LLMs and constructing the corresponding counter-memory. We reuse MA and CMA generated by \citet{xie2024knowledgeconflict} for our experiments.
    
    \item \textbf{Natural Questions} \cite{47761} is widely used in open-domain QA research. It consists of manually crafted questions based on selected paragraphs from Wikipedia, and the subjects in questions of the NQ dataset are generally more popular and commonly known. 
    \citet{longpre2021entity} provide a test set that is used to test the context faithfulness of LLMs by substituting entity of the NQ dataset. The entity substitute involves five categories: person (PER), date (DAT), numeric (NUM), organization (ORG), and location (LOC). The test set contains 4,685 samples, including 1,667 unique questions.
\end{itemize}

\subsection{Memory Strength}\label{sec:memory strength}

Inspired by \citet{zhao-etal-2024-knowing}, we use the consistency of answers to different paraphrases of the same question $Q$ to measure the LLM's memory strength $S_Q$ for the knowledge $K_Q$ associated with the question. This method is motivated by the intuition that if an LLM does not have a strong memory of a question, 
it will often give different answers when asked with paraphrased questions that are semantically equivalent, 
as shown in Table \ref{tab:qa}. In contrast, it can produce consistent answers if the LLM has a strong memory of a question. The process involves two key steps: First, several paraphrased versions of the original question are generated with ChatGPT\footnote{https://platform.openai.com/docs/models/gpt-3-5-turbo, the specific version is 0125.}, and the answers to those paraphrased questions are clustered (Section \ref{section:question_paraphrase}). Then, memory strength $S_Q$ is calculated using answer consistency (Section \ref{method:calculating}).

\subsubsection{Question Paraphrases and Answer Clustering}\label{section:question_paraphrase}

The prompt used for paraphrasing the question is provided in Tabel \ref{tab:prompts} (index 1) in the Appendix. 
For each question $Q$, we generate $n$ paraphrases $\{P_1, \cdots, P_n\}_Q$. For the NQ dataset, we paraphrase the question in each data sample directly. For the popQA dataset, we paraphrase the question template for each relation type since all questions of the same relation type share the same question template. To ensure the paraphrased questions are proper to use, we check if two paraphrased questions are semantically equivalent with an LLM\footnote{\label{llama3.1}https://huggingface.co/meta-llama/Meta-Llama-3.1-8B}. The prompt for this semantic equivalence detection is provided in Table \ref{tab:prompts} (index 2). For any paraphrase that is deemed not equivalent, we ask the LLM to re-generate it until a satisfactory version is produced.

Next, LLMs answer the paraphrased questions $\{P_1, \cdots, P_n\}_Q$ in a closed-book setting. We denote the answers as $\{A_1, \cdots, A_n\}_Q$. The answers are grouped into several clusters based on their consistency. The clustering is done by checking answers incrementally. If an answer matches any answer within an existing cluster, this answer is added to this cluster; if not, a new cluster is created with this answer. We use an LLM\textsuperscript{\ref{llama3.1}} to determine whether two answers are consistent. The prompt used for this answer inconsistency detection is shown in Table \ref{tab:prompts} (index 3). We denote the clusters for question $Q$ as $\{c_1, \cdots, c_m \}_Q$. 


\subsubsection{Calculating Memory Strength}\label{method:calculating}



Once answer clusters $\{c_1, \cdots, c_m \}_Q$ are identified, memory strength $S(Q)$ can be obtained by calculating the negative entropy of cluster distribution. The formula is 
\begin{equation}
    S(Q) = \sum_{i=1}^m \frac{N(c_i)}{n} log \frac{N(c_i)}{n},
\end{equation}
where $N(c_i)$ is the number of answers in the cluster $c_i$, and $n$ is the number of paraphrases for question $Q$. The memory strength score is a non-positive value. A larger score indicates a stronger memory (0 is the maximum value of memory strength score). In the experiments, we set $n=7$ for all the questions in the NQ and popQA datasets. Memory strength reflects how well the LLM remembers the required knowledge: the weaker the memory, the more random and inconsistent the answers are.

\subsection{MA, CMA, and Evidence Generation}

\subsubsection{MA and CMA Generation}

For the popQA dataset, both MA and CMA are obtained following the method described in \citet{xie2024knowledgeconflict}. For the MA of the NQ dataset, we also use a closed-book approach, similar to \citet{xie2024knowledgeconflict}. While, the process of generating CMA differs from the process of generating CMA in \citet{xie2024knowledgeconflict}. 
Unlike the popQA dataset, the NQ dataset does not provide relation types for the questions or offer sets of subject and object entities for substitution.
To address this issue, we propose an approach using an LLM to substitute entities in MA to generate CMA. First, we identify which ``wh-'' question type\footnote{which refers to what, when, where, who, whom, which, whose, why, and how.} the question belongs to using string matching. Then, based on the question type, we determine the type of entity to be replaced in the MA. Finally, we use an LLM to make the substitution. For example, in Figure \ref{fig:framework}, the question ``how many episodes...'' is of the type ``how\_many'', so the entity to be replaced in the MA ``there are 23 episodes...'' should be a \textit{NUMBER}. We let ChatGPT perform the substitution with an alternative entity. The prompt used is shown in Table \ref{tab:prompts} (index 5). We have the detailed description for generating CMA in Appendix \ref{sec:CMA generation}.

\textbf{CMA filter}. As noted in Section \ref{sec:memory strength}, LLMs can produce multiple MAs for the same question. To ensure the CMA conflicts with MAs, we require that the CMA is different from any of the answers $\{A_1, \cdots, A_n\}$ generated in Section \ref{section:question_paraphrase}, so the alternative entity should not appear in MAs. For the popQA dataset, the alternative entity is known. For the NQ dataset, we first identify the alternative entity in the CMA by comparing the MA and CMA, and then check if this entity appears in any of the MAs $\{A_1, \cdots, A_n\}$. We filter out data samples whose CMA does not conflict with MAs.

\subsubsection{Evidence Generation}

In this section, we explain how to generate different styles of evidence. We classify evidence into two categories: direct evidence and indirect evidence.

\textbf{Direct evidence} is a semantically equivalent statement of the CMA, providing the clearest support for the claim made by the CMA. We generate the direct evidence by paraphrasing the CMA with ChatGPT, following the prompt shown in Table \ref{tab:prompts} (index 6). For example, in Figure \ref{fig:framework}, the CMA ``there are 15 episodes in Chicago Fire season 4'' is paraphrased to ``season 4 of Chicago Fire consists of a total of 15 episodes''. These two statements are semantically equivalent. 

To ensure the reliability of the evidence, the evidence must mutually entail with the CMA. This entailment is verified using an NLI model \footnote{\label{nli_model}https://huggingface.co/microsoft/deberta-v2-xxlarge-mnli}. 
Direct evidence is intuitive, simple, and coherent, making it the straightforward type of evidence for the LLM to process. If the LLM is receptive to external evidence, it should be able to adopt direct evidence easily.



\textbf{Indirect evidence} differs from direct evidence by adding extra details that provide a more thorough description of the subject related to the CMA. This additional information makes the evidence more comprehensive and might be more persuasive. For example, in Figure \ref{fig:framework}, the indirect evidence includes details not found in the counter answer, such as the title of the first episode and its release date, along with the fact that there are 15 episodes in total. The prompt to generate indirect evidence is shown in Table \ref{tab:prompts} (index 7).

To ensure the reliability of indirect evidence, the indirect evidence should entail the CMA and the additional information introduced by the evidence should not entail the MA. Otherwise, the indirect evidence can support both the MA and CMA. The NLI model\textsuperscript{\ref{nli_model}} is used to verify that indirect evidence entailed with CMA and neutral or contradictory with MA.

For both direct and indirect evidence, if the content generated by the LLM does not meet the required conditions, we prompt the LLM to regenerate it up to five times. If it still fails after five attempts, we exclude that question from the dataset.

\section{Experiments}






In this study, we aim to investigate two key research questions. 1) Does memory strength have an impact on the context faithfulness of LLMs? 2) Does the style of evidence affect the context faithfulness of LLMs? These research questions are explored in Section \ref{sec:impact_memory_strength} and \ref{sec:impact_evidence_style}, respectively. We also provide additional studies in Appendix \ref{sec:additional_studies}, which includes a study about the impact of option order and a case study.




\subsection{Experiment Setup}

\textbf{LLM Models.} Our experiments are conducted using six well-known language models: ChatGPT \cite{openai2022chatgpt}, GPT-4 \cite{openai2023gpt4}, LLaMA2-7B, LLaMA2-70B \cite{touvron2023llama2openfoundation}, LLaMA3.2-3B \cite{Meta2024LLaMA}, and Claude3.5 \cite{anthropic2024claude}. These models represent a diverse range of architectures and capabilities. ChatGPT, GPT-4, and Claude3.5 are cutting-edge models developed by OpenAI and Anthropic. LLaMA2, with its 7 billion and 70 billion parameter variants, is a strong open-source alternative that has demonstrated competitive performance across a wide variety of tasks. LLaMA3.2-3B is the newest version of LLaMA. The inclusion of models with varying scales (from 3B to 70B) and training methodologies allows us to explore both closed-source systems (GPT-4, ChatGPT, and Claude3.5) and open-source solutions (LLaMA2-7B, LLaMA-70B and LLaMA3.2-3B).


\noindent
\textbf{Evaluation Metrics.} Following previous work \cite{longpre2021entity, xie2024knowledgeconflict, chen-etal-2022-rich}, we transform the short answer QA to a multiple-choice QA format by providing a few options as possible answers\footnote{\citet{xie2024knowledgeconflict} shows that answer consistency between short answer and multi-choice are 94\%, 96\% and 92\% for ChatGPT, GPT-4 and LLaMA2-7B, respectively.}. This limits the answer generation space and makes it easy to evaluate without manual checking. Specifically, for each question from both datasets, LLMs are instructed to select one
answer from the MA, CMA, and “Uncertain” (UCT). We report the ratio of MA ($R_{m}$), CMA ($R_{c}$), and UCT ($R_{u}$) as calculated below:
\begin{align}
    R_{m} &= \frac{f_m}{f_m + f_c + f_u}  \nonumber \\
    R_{c} &= \frac{f_c}{f_m + f_c + f_u}  \nonumber \\
    R_{u} &= \frac{f_u}{f_m + f_c + f_u},
\end{align}
where $f_m, f_c$, and $f_u$ are the count of questions with MA, CMA, and UCT answers, respectively.

\subsection{Role of Memory Strength}\label{sec:impact_memory_strength}



\begin{figure*}[t]
    \centering
    \includegraphics[width=\linewidth]{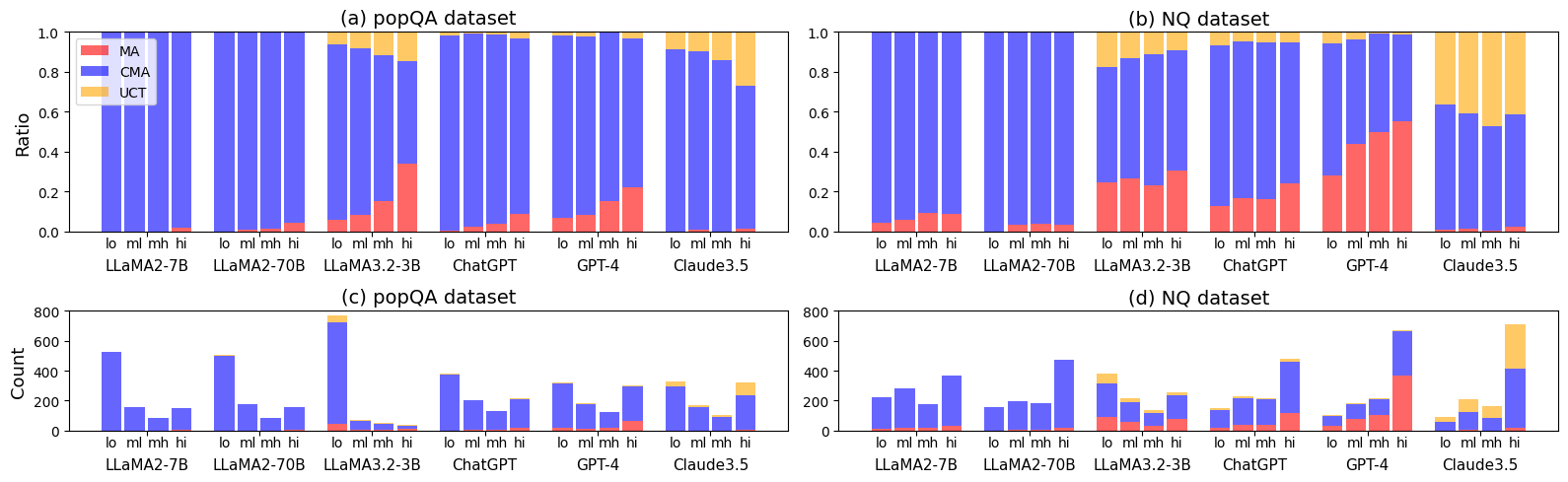}
    \caption{Relationship between Memory Strength and Different Answers' Ratios across popQA and NQ Datasets (with Direct Evidence).
The figure presents the ratio and count of MA, CMA, and UCT across four memory strength groups: low(lo), mid-low(ml), mid-high(mh), and high(hi). The results show a clear positive correlation between memory strength and MA ratio($R_m$).}
    \label{fig:memory_strength_memory_ratio}
\end{figure*}

\subsubsection{Correlation between Context Faithfulness and Memory Strength}\label{sec:faith_with_memory_strength}
To demonstrate the relationship between context faithfulness and memory strength, we categorize the questions in each dataset into four groups according to the memory strength of each LLM. The four groups are low, mid-low, mid-high, and high, corresponding to the memory strength intervals [-2, -1], (-1, -0.5], (-0.5, -0.25] and [-0.25, 0], respectively. We use the direct evidence in this experiment. The results are shown in Figure \ref{fig:memory_strength_memory_ratio}\footnote{We put results for other evidence styles in Appendix, Figures \ref{fig:memory_strength_memory_ratio_dual_paraphrase}, \ref{fig:memory_strength_memory_ratio_two_sentence}, and \ref{fig:memory_strength_memory_ratio_dircet_counter}. The conclusion is consistent.}. Figure \ref{fig:memory_strength_memory_ratio} (a)(b) shows the ratios of questions with MA, CMA, and UCT answers for the NQ and popQA datasets, respectively. Note that different LLMs may have different memory strengths to the same question. Therefore, both the specific questions and the count of questions in the same group can vary across different LLMs. To illustrate this, we present the count of questions in each group (low, mid-low, mid-high, and high) in Figure \ref{fig:memory_strength_memory_ratio} (c)(d) for popQA and NQ datasets, respectively. We can draw the following conclusions.


\textbf{There is a clear positive correlation between memory strength and MA ratio for individual LLMs}. From Figure \ref{fig:memory_strength_memory_ratio} (a)(b), it is obvious that for all tested LLMs, the ratio of MA (red) increases when memory strength increases, while the ratio of CMA (blue) decreases.  This trend is also consistent across both datasets. It is more obvious for GPT models. Among the tested LLMs, Claude3.5 tends to choose UCT options more often, especially on questions with high memory strength.


\textbf{Memory Strength Increases with Model Scale.} 
We can observe from Figure \ref{fig:memory_strength_memory_ratio} (c)(d) that larger LLMs, such as GPT-4, have more data samples in the high memory strength group (hi), while smaller LLMs, such as LLaMA3.2-3B, have more samples in the low memory strength group (lo). This aligns with the common intuition that larger LLMs, with more parameters, have more knowledge than smaller LLMs. Further evidence and discussion can be found in the Appendix \ref{appendix: memory strength on datasets}.


\subsubsection{LLMs Performance Analysis}

\begin{figure*}[t]
    \centering
    \includegraphics[width=0.96\linewidth]{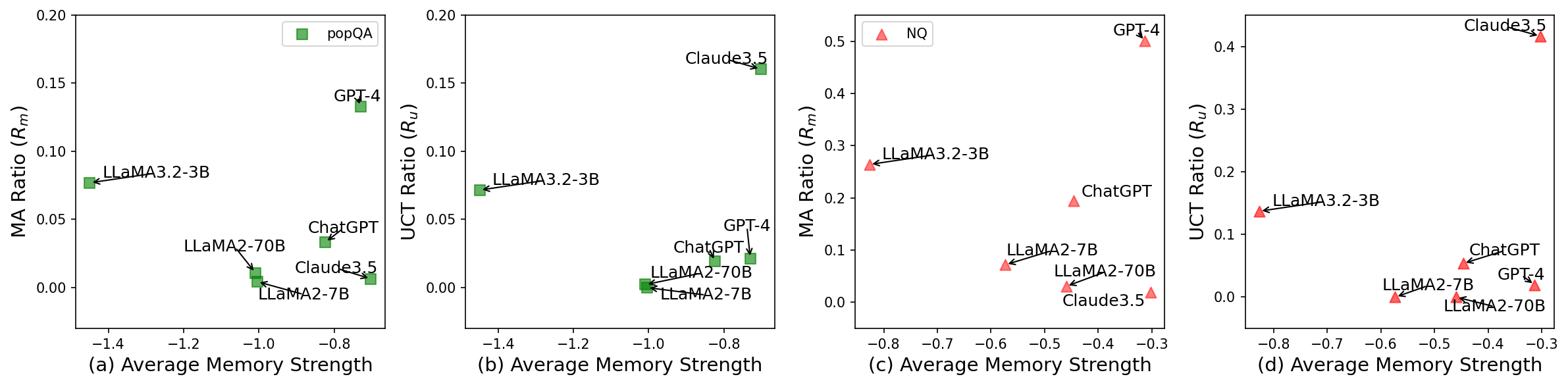}
    \caption{MA Ratio ($R_m$) and UCT Ratio ($R_u$) VS Memory Strength across PopQA and NQ Datasets. The figure presents the relationship between average memory strength and the MA ratio as well as the UCT ratio for all six LLMs. It shows that lower average memory strength does not always mean better context faithfulness for different LLMs and newer versions of GPT and LLaMA models seem to ignore context faithfulness issues.}
    \label{fig:llm_compare}
\end{figure*}



In the aforementioned conclusion, different LLMs have different knowledge. Thus, testing LLMs' context faithfulness on the same dataset may be unfair.
To illustrate this, we compute the average memory strength of LLMs on PopQA and NQ datasets, respectively, along with their MA and UCT ratios. The results are presented in Figure \ref{fig:llm_compare}. We can draw the following conclusions.

First, for different LLMs, a lower average memory strength does not necessarily imply better context faithfulness. For example, Claude 3.5 has a high average memory strength, implying that Claude 3.5 is a knowledgeable LLM, but it has a low MA ratio ($R_m$) and a high UCT ratio ($R_u$). This indicates that Claude 3.5 tends to refuse to answer when facing knowledge conflicts. In contrast, LLaMA3.2-3B, despite having much less knowledge (low average memory strength), has the second-highest MA ratio ($R_m$). This means that LLaMA3.2-3B relies heavily on limited internal knowledge when facing knowledge conflicts, implying that it is not context-faithful. 

Second, newer versions of GPT and LLaMA models appear to overlook context faithfulness issues during the training process. GPT-4, with slightly more knowledge than Chatgpt, shows a significantly higher MA ratio ($R_m$), and LLaMA3.2-3B, with the least knowledge, shows a higher MA ratio ($R_m$) than the two LLaMA2 models. 

Finally, a new context faithfulness evaluation metric is needed. This metric should consider both the different answers' ratios and memory strength when LLMs encounter knowledge conflicts. Simply using the MA ratio (which is used widely to measure context faithfulness currently) to evaluate context faithfulness across different LLMs may not be fair and sufficient.

\subsection{Role of Evidence Style}\label{sec:impact_evidence_style}

\begin{table*}[t]
\centering
\resizebox{\textwidth}{!}{
\begin{tabular}{l|c|c|c|c|c|c|c|c|c|c|c|c}
\hline
\textbf{} & \multicolumn{6}{c|}{\textbf{popQA}} & \multicolumn{6}{c}{\textbf{NQ}} \\ \cline{2-13}
\textbf{} & \textbf{LLaMA2-7B} & \textbf{LLaMA2-70B} & \textbf{LLaMA3.2-3B} & \textbf{ChatGPT} & \textbf{GPT-4} & \textbf{Claude3.5} & \textbf{LLaMA2-7B} & \textbf{LLaMA2-70B} & \textbf{LLaMA3.2-3B} & \textbf{ChatGPT} & \textbf{GPT-4} & \textbf{Claude3.5} \\ \hline
\textbf{$\#$ of Q (Initial)} & 1000 & 1000 & 1000 & 1000 & 1000 & 1000 & 1667 & 1667 & 1667 & 1667 & 1667 & 1667 \\ \hline
\textbf{$\#$ of Q with direct evidence} & 918 & 922 & 938 & 933 & 933 & 931 & 1042 & 1009 & 1002 & 1079 & 1171 & 1173\\ \hline
\textbf{$\#$ of Q with indirect evidence} & 901 & 895 & 917 & 911 & 918 & 913 & 976 & 972 & 941 & 1025 & 1108 & 1059 \\ \hline
\end{tabular}
}
\caption{Number of final examples for each LLM. The difference between LLMs is due to their different outputs going through the framework.}
\label{tab:final_count}
\end{table*}

\begin{table*}[t]
\centering
\resizebox{\textwidth}{!}{%
\begin{tabular}{c|l|c|ccc|ccc|ccc|ccc|ccc|ccc}
\hline
    \multirow{2}{*}{Dataset} & \multirow{2}{*}{Evidence Style} & \multirow{2}{*}{S \#}  & \multicolumn{3}{c}{LLaMA2-7B} & \multicolumn{3}{c}{LLaMA2-70B} & \multicolumn{3}{c}{LLaMA3.2-3B} & \multicolumn{3}{c}{ChatGPT} & \multicolumn{3}{c}{GPT-4}  & \multicolumn{3}{c}{Claude3.5}   \\ 
    \cline{4-21}
              &  & & $R_m\downarrow$ & $R_c\uparrow$ & $R_u$ & $R_m\downarrow$ & $R_c\uparrow$ & $R_u$ & $R_m\downarrow$ & $R_c\uparrow$ & $R_u$ & $R_m\downarrow$ & $R_c\uparrow$ & $R_u$ & $R_m\downarrow$ & $R_c\uparrow$ & $R_u$ & $R_m\downarrow$ & $R_c\uparrow$ & $R_u$ \\ \hline
             & \multicolumn{20}{c}{\cellcolor[gray]{0.9} Group 1: Q with direct evidence} \\ \cline{2-21}
\multirow{11}{*}{popQA} & \multirow{3}{*}{Direct Evidence} & 1 & 0.44 & 99.56 & 0.0 & 1.08 & 98.7 & 0.22 & 7.69 & 85.15 & 7.16 & 3.32 & 94.75 & 1.93 & 13.29 & 84.57 & 2.14 & 0.65 & 83.31 & 16.04  \\ \cline{3-21}
&                                   & 2 &  0.44 & 99.56 & 0.0 & 0.98 & 98.81 & 0.22 & 2.67 & 95.83 & 1.5 & 2.79 & 96.03 & 1.18 & 4.39 & 93.46 & 2.14 & 0.32 & 94.08 & 5.6  \\ \cline{3-21}
&                                   & 3 &  0.65 & 99.35 & 0.0 & 1.08 & 98.7 & 0.22 & 2.56 & 94.98 & 2.46 & 3.0 & 95.5 & 1.5 & \underline{2.57} & 95.28 & 2.14 & \underline{0.21} & 94.73 & 5.06
  \\ \cline{2-21}
& \multirow{2}{*}{Direct+Paraphrase} & 2 & 0.22 & 99.78 & 0.0 & 1.19 & 98.7 & 0.11 & 3.1 & 95.62 & 1.28 & 2.36 & 97.0 & 0.64 & 3.0 & \underline{95.71} & 1.29 & \underline{0.21} & 95.69 & 4.09  \\ \cline{3-21}
&                                   & 3 & 0.11 & 99.89 & 0.0 & 0.43 & 99.35 & 0.22 & \textbf{1.07} & \underline{97.86} & 1.07 & \textbf{1.39} & \textbf{98.28} & 0.32 & \textbf{1.29} & \textbf{98.5} & 0.21 & \textbf{0.11} & \textbf{98.06} & 1.83  \\ \cline{2-21}
& \multicolumn{20}{c}{\cellcolor[gray]{0.9} Group 2: Q with indirect evidence} \\ \cline{2-21}
& Direct Evidence                           & 1 & 0.44 & 99.56 & 0.0 & 0.45 & 99.33 & 0.22 & 7.31 & 85.61 & 7.09 & 3.07 & 95.28 & 1.65 & 12.53 & 85.29 & 2.18 & 0.66 & 83.9 & 15.44  \\ \cline{2-21}
& \multirow{2}{*}{Indirect Evidence}         & 2 & \underline{0.0} & \underline{100.0} & 0.0 & \underline{0.11} & \underline{99.89} & 0.0 & 4.36 & 89.64 & 6.0 & 3.18 & 96.38 & 0.44 & 13.51 & 85.73 & 0.76 & 0.88 & 82.37 & 16.76  \\ \cline{3-21}
&                                   & 3 & \textbf{0.0} & \textbf{100.0} & 0.0 & \textbf{0.0} & \textbf{100.0} & 0.0 & 3.27 & 92.58 & 4.14 & 1.76 & 97.91 & 0.33 & 9.26 & 90.2 & 0.55 & 0.88 & 87.95 & 11.17   \\ \cline{2-21}
& \multirow{2}{*}{Direct+Indirect} & 2 & 0.22 & 99.78 & 0.0 & \underline{0.11} & 99.78 & 0.11 & 3.71 & 93.24 & 3.05 & 1.87 & 97.26 & 0.88 & 7.95 & 91.18 & 0.87 & 0.88 & 93.43 & 5.7    \\ \cline{3-21} 
&                                   & 3 & 0.11 & 99.89 & 0.0 & \underline{0.11} & 99.78 & 0.11 & \underline{1.42} & \textbf{98.15} & 0.44 & \underline{1.43} & \underline{98.13} & 0.44 & 5.12 & 94.77 & 0.11 & 0.88 & \underline{96.28} & 2.85  \\ \hline
             &  \multicolumn{20}{c}{\cellcolor[gray]{0.9} Group 1: Q with direct evidence} \\  \cline{2-21}
\multirow{11}{*}{NQ} & \multirow{3}{*}{Direct Evidence} & 1 & 7.2 & 92.8 & 0.0 & 3.07 & 96.93 & 0.0 & 26.41 & 59.88 & 13.71 & 19.46 & 75.16 & 5.38 & 50.04 & 47.99 & 1.96 & 1.96 & 56.4 & 41.64 \\ \cline{3-21}
&                                   & 2 &  5.47 & 94.53 & 0.0 & 3.07 & 96.93 & 0.0 & 16.93 & 75.2 & 7.86 & 19.09 & 76.83 & 4.08 & 20.24 & 77.54 & 2.22 & 0.77 & 83.7 & 15.53 \\ \cline{3-21}
&                                   & 3 &  6.81 & 93.19 & 0.0 & 2.68 & 97.22 & 0.1 & 11.79 & 76.11 & 12.1 & 22.06 & 72.75 & 5.19 & \underline{17.34} & \underline{80.87} & 1.79 & \textbf{0.26} & 88.22 & 11.52  \\ \cline{2-21}
& \multirow{2}{*}{Direct+Paraphrase} & 2 & 4.13 & \underline{95.87} & 0.0 & 1.49 & 98.41 & 0.1 & 13.21 & 79.33 & 7.46 & 15.29 & 81.28 & 3.43 & 18.96 & 79.67 & 1.37 & \underline{0.34} & \underline{88.65} & 11.01 \\ \cline{3-21}
&                                   & 3 & \textbf{3.26} & \textbf{96.74} & 0.0 & \textbf{1.19} & \textbf{98.61} & 0.2 & \underline{9.38} & \underline{83.27} & 7.36 & \underline{9.55} & \underline{86.75} & 3.71 & \textbf{11.27} & \textbf{87.28} & 1.45 & \textbf{0.26} & \textbf{93.09} & 6.65 \\ \cline{2-21} 
& \multicolumn{20}{c}{\cellcolor[gray]{0.9} Group 2: Q with indirect evidence} \\ \cline{2-21}
& Direct Evidence                      & 1 & 5.53 & 94.47 & 0.0 & 2.67 & 97.32 & 0.0 & 23.19 & 63.73 & 13.08 & 18.73 & 75.71 & 5.56 & 48.83 & 49.19 & 1.99 & 1.77 & 59.44 & 38.79  \\ \cline{2-21}
& \multirow{2}{*}{Indirect Evidence}               & 2 & \underline{3.28} & 95.29 & 1.43 & 1.65 & 98.25 & 0.1 & 13.32 & 77.77 & 8.92 & 13.66 & 84.1 & 2.24 & 44.59 & 53.7 & 1.71 & 1.98 & 67.88 & 30.14  \\ \cline{3-21}
&                                   & 3 & 4.82 & 94.06 & 1.13 & 1.85 & 97.84 & 0.31 & 11.53 & 80.86 & 7.61 & 13.27 & 84.19 & 2.54 & 39.89 & 58.57 & 1.53 & 3.65 & 71.32 & 25.03 \\ \cline{2-21}
& \multirow{2}{*}{Direct+Indirect}               & 2 & 5.33 & 94.67 & 0.0 & \underline{1.34} & 98.25 & 0.41 & 9.75 & 82.28 & 7.97 & 12.68 & 84.78 & 2.54 & 32.4 & 66.06 & 1.53 & 1.36 & 80.92 & 17.73 \\ \cline{3-21}
&                                   & 3 & 4.41 & 95.59 & 0.0 & 1.44 & \underline{98.56} & 0.0 & \textbf{8.32} & \textbf{85.49} & 6.18 & \textbf{9.46} & \textbf{88.1} & 2.44 & 28.7 & 69.67 & 1.62 & 1.77 & 83.94 & 14.29  \\ \hline

\end{tabular}
}
\caption{Results of LLM Receptiveness to Different Evidence Styles Across NQ and popQA Datasets.
The table presents the MA ratio ($R_m$), CMA ratio ($R_c$), and uncertain answer ratio ($R_u$) for various evidence styles across six models. All the ratios are in \%. The best results are highlighted in \textbf{bold}, and the second-best results are \underline{underlined}.}
\label{tab:main_table}
\end{table*}

\noindent
\textbf{Evidence Styles.} 
We formulate four types of evidence styles:
\textbf{1)} Direct Evidence. This is the most straightforward form of evidence and serves as our baseline. To assess the impact of evidence length, we also create versions where the direct evidence is repeated twice and three times for comparison. 
\textbf{2)} Direct Evidence Combined with Paraphrases of CMA. To examine the effect of evidence phrasing and expression, we combine the direct evidence with one paraphrase of the CMA to form a two-sentence evidence and with two paraphrases to form a three-sentence evidence. 
\textbf{3)} Indirect Evidence. We generate indirect evidence consisting of two sentences and three sentences, respectively\footnote{We regulate the length of the generated evidence to control the influence of evidence length. The prompts used are detailed in Table \ref{tab:prompts} (index 7) in Appendix.}.
\textbf{4)} Direct Evidence Combined with Indirect Evidence. We combine the direct evidence with the first sentence of the two-sentence indirect evidence to form a two-sentence evidence and with both sentences to form a three-sentence evidence. 


Table \ref{tab:final_count} presents the final number of instances used for evaluation. We observe a slight difference in the quantities of questions with direct and indirect evidence since it is easier for ChatGPT to generate direct evidence that meets our requirements. The specific number of instances at each step in evidence generation is detailed in Table \ref{tab:evidence_count} in the Appendix. Due to the quantity difference between direct evidence and indirect evidence, we divide the styles of evidence into two groups: Group 1 includes Direct Evidence and Direct + Paraphrase evidence. Group 2 includes Indirect Evidence and
Direct + Indirect evidence. Each group has different Direct Evidence results serving as baselines.



Table \ref{tab:main_table} shows the results of different evidence styles. We can make the following observations and conclusions. 

\textbf{In Group 2, the MA Ratio ($R_m$) of direct evidence is slightly lower than that in Group 1}. 
During the evidence generation, there are some questions for which ChatGPT can provide direct evidence but cannot produce indirect evidence. Removing these questions leads to a decrease in $R_m$ of direct evidence with one sentence, which implies that LLMs have a relatively high $R_m$ for the removed questions. But in general, the $R_m$ of direct evidence with one sentence in Group 1 is close to that in Group 2, so the results of Group 1 and Group 2 are \textbf{still comparable}.

\textbf{Simple repetition of direct evidence is not always effective}. Comparing direct evidence with one to three sentences, we observe similar $R_m$ and $R_c$ for LLaMA2-7B, LLaMA2-70B, and ChatGPT. For LLaMA3.2-3B, GPT-4, and Claude3.5, $R_m$ of direct evidence with two and three sentences decreases significantly. The results imply that the newer LLMs are receptive to evidence repeated multiple times.


\textbf{Paraphrasing direct evidence is highly effective across all models and datasets}. Comparing Direct Evidence with two and three sentences and Direct + Paraphrase with two and three sentences, respectively, we observe $R_m$ of the latter significantly decreases. For example, $R_m$ is reduced by more than half, comparing Direct + Paraphrase with three sentences with Direct Evidence with three sentences on the popQA dataset for all tested LLMs. The result implies that paraphrasing is an effective method to enhance the receptiveness of LLMs to external evidence.


\textbf{Indirect Evidence improves LLMs' receptiveness to CMAs, but less effectively than paraphrasing}. Comparing Indirect Evidence with two and three sentences with Direct Evidence with one sentence, $R_m$ decreases for almost all LLMs, but the reduction is not significant compared to the Direct + Paraphrase evidence with two or three sentences. It implies that adding detailed information is less effective than paraphrasing direct evidence.


\textbf{Combining Direct evidence with Indirect evidence generally enhances persuasiveness}. Comparing Direct + Indirect evidence with Indirect Evidence, $R_m$ decreases except for LLaMA2-7B. For example, comparing Direct + Indirect with three sentences and Indirect Evidence with three sentences, $R_m$ has an obvious decrease. The result implies that adding direct evidence to indirect evidence is effective in improving LLMs' receptiveness to CMAs.

\section{Conclusion}

We investigate how context-faithful LLMs are to external evidence across two datasets, PopQA and NQ datasets, using LLaMA2-7B, LLaMA2-70B, ChatGPT, GPT-4, LLaMA3.2-3B and Claude3.5. Our findings highlight the critical role of memory strength in shaping LLM behavior. There is a clear positive correlation between memory strength and memory answer ratio. Furthermore, we demonstrate that paraphrasing significantly enhances the context faithfulness of LLMs across various models and datasets. 
These findings offer valuable insights for advancing research in retrieval-augmented generation and context-based LLM applications.


\section*{Limitations}

Our framework does not process all types of questions in the NQ dataset. Although it effectively handles the majority of NQ questions, it currently lacks the capability to address "what," "how," and "why" question types. The omission of these questions may introduce some bias into our results. Similar to previous studies, our study also focuses on knowledge conflict for extractive QA tasks, where the answer must appear in the evidence. Our conclusion may not be extendable to other types of QA tasks, such as abstractive QA and generative QA.

We employed a Natural Language Inference (NLI) model to detect and filter the generated data. Although the NLI model demonstrates high accuracy and the quality of generated data is high, it still cannot guarantee complete correctness. Further, since the NLI model is also trained using language models, which may be biased with parametric memory, it may introduce biases facing knowledge conflicts. 


\section*{Acknowledgments}
The work was supported in part by the US National
Science Foundation under grant NSF-CAREER 2237831.
We also want to thank the anonymous reviewers for their helpful comments.

\clearpage
\bibliography{custom}

\clearpage
\appendix

\section*{Appendix}\label{sec:appendix}

Within this supplementary material, we elaborate
on the following aspects:

\begin{itemize}
    \item \ref{sec:methods details} Methodology Details
    \item \ref{sec:additional_studies} Additional Studies
    \item \ref{sec:prompts} Prompts
\end{itemize}

\section{Methodology and Experiment Details}\label{sec:methods details}

\subsection{CMA Generation for NQ dataset}\label{sec:CMA generation}


Departing from prior approaches \cite{longpre2021entity} that use NER techniques \cite{chen-etal-2021-evaluating, zhou-etal-2022-distantly,li-etal-2025-examine} to locate and substitute answer entities, we employ LLMs to automatically generate alternative entities. Our CMA construction process from MA consists of three steps: 1) identify question type, 2) determine entity type in MA to change, and 3) generate CMA with LLMs.

\begin{figure}[t]
    \centering
    \includegraphics[width=0.82\linewidth]{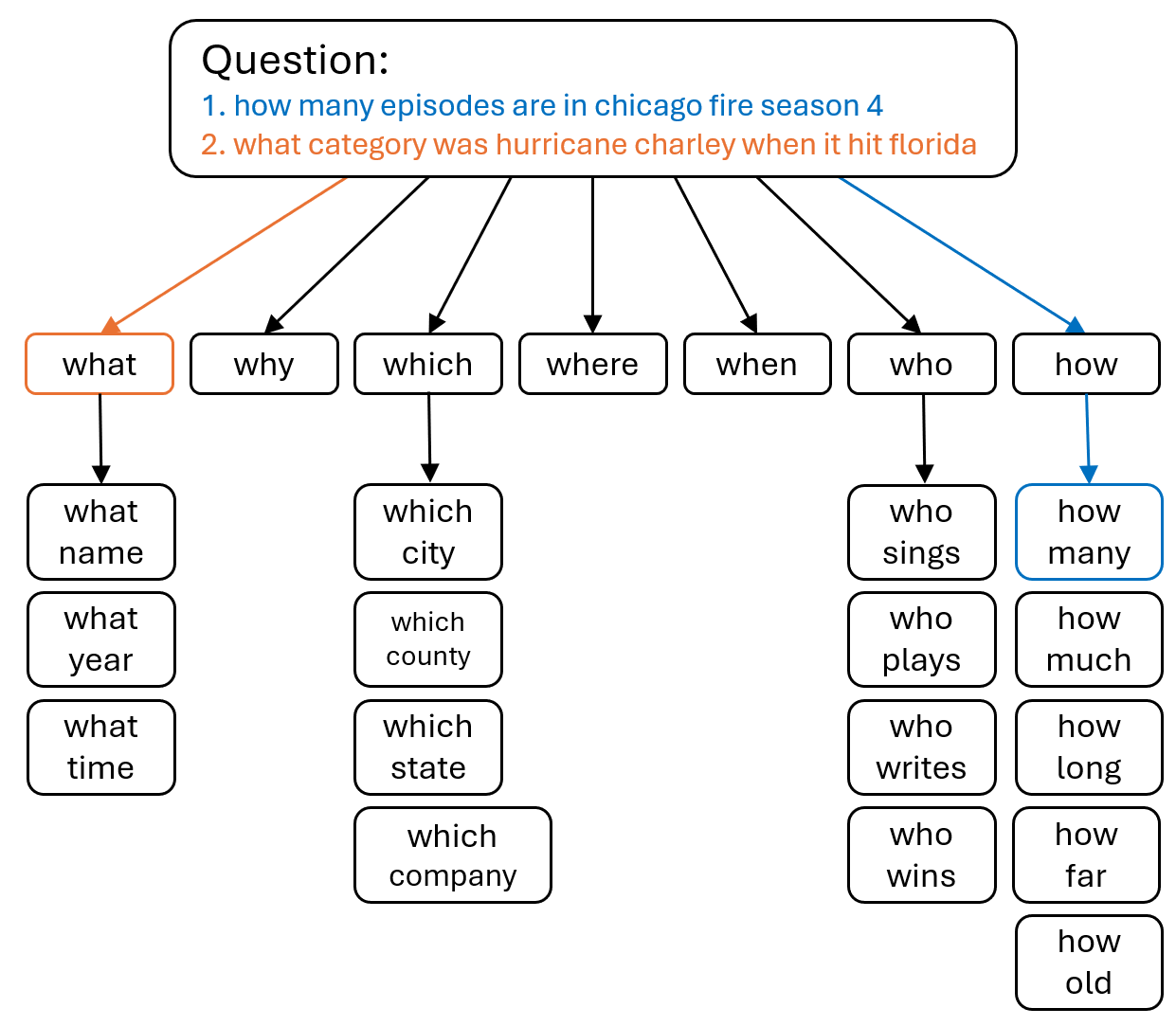}
    \caption{Two-layer Question Typing Tree}
    \label{fig:type_tree}
\end{figure}

\textbf{Identity Question Type:} We first build a typing tree using rules to categorize questions. Figure \ref{fig:type_tree} illustrates the typing tree, which consists of a two-layer structure. In the typing process, we first determine if a question begins with one of the following words: ``what'', ``when'', ``where'', ``which'', ``who'', ``why'', or ``how''. 
However, this approach can still group different types of questions together. To address this, we use a second layer to refine the typing by analyzing two specific words in the question. For example, the question shown in Figure \ref{fig:framework} falls into the ``how\_many'' category. Table \ref{tab:question_types} shows the statistics of question types of the NQ dataset. Note that, we find 127 samples that are not questions in the process, so we list them as ``other''.

\textbf{Determine Entity Type in MA to Change:} After categorizing the questions, we determine the entity type in MA needs to be replaced. To achieve this, we give each type of question an entity type, and many questions can share the same entity type. For example, both ``when'' and ``what year''  ask for a time. So a time entity in MA should be substituted. The final set of entity types is summarized in Table \ref{tab:keywords}. We do not process questions starting with ``what'', ``which'' or ``how'' due to the lack of a unified entity type for these questions. Table \ref{tab:exclude_quesition_types} shows the statistics of the unprocessed questions. 
These question types account for only 7.3\% of the NQ dataset (with 2, 98, 22, and 0 instances for ``how'', ``what'', ``which'', and ``why'', respectively), and this issue does not occur in the PopQA dataset. Since the overall trends are consistent across both datasets, we believe the missing questions will not affect our conclusions.

\begin{table}[t]
\centering
\resizebox{0.182\textwidth}{!}{
\begin{tabular}{|l|r|}
\hline
\textbf{Question Type} & \textbf{Count} \\
\hline
how\_many & 97 \\
how\_much & 1 \\
how\_long & 3 \\
how\_old & 3 \\
how & 2 \\
who\_sings & 100 \\
who\_plays & 179 \\
who\_writes & 65 \\
who\_wins & 55 \\
who & 479 \\
where & 138 \\
when & 276 \\
what\_year & 7 \\
what\_name & 4 \\
what & 98 \\
which\_country & 6 \\
which\_city & 2 \\
which\_state & 2 \\
which\_year & 1 \\
which & 22 \\
why & 0 \\
other & 127 \\ \hline
total & 1667 \\
\hline
\end{tabular}
}
\caption{Distribution of Question Types and their Counts}
\label{tab:question_types}
\end{table}

\textbf{Generate CMA with LLMs:} Instead of manually editing the MA, we leverage the LLMs' ability to generate CMA by providing it with a carefully designed prompt, which is shown in Table \ref{tab:prompts} (index 5). This prompt instructs the LLM to replace the entity with a certain type in the MA (from Step 2) with an alternative, ensuring the generated CMA differs from the MA. 

The generated CMA must meet two key criteria: 1) The CMA must directly contradict the MA. To ensure this, we employ a Natural Language Inference (NLI) model\footnote{https://huggingface.co/microsoft/deberta-v2-xxlarge-mnli} to verify the contradiction between the two answers. 2) The alternative entity in CMA must not appear in the question. We achieve this check by string matching. 
If the CMA fails to meet either of these criteria, we prompt the LLM to regenerate the CMA up to 5 times. If no proper CMA is generated, we filter out this question. 

\begin{table}[t]
\centering
\resizebox{0.48\textwidth}{!}{%
\begin{tabular}{l|l}
\hline
\textbf{Question Type} & \textbf{Key Term} \\ \hline
when, what\_year, which\_year, how\_long & time \\ \hline
where, which\_city, which\_state, which\_country & location \\ \hline
who, what\_name & name of person \\ \hline
how\_many, how\_much & number \\ \hline
who\_sings & singer’s name \\ \hline
who\_plays & player’s name \\ \hline
who\_writes & writer’s name \\ \hline
who\_wins & winner’s name \\ \hline
how\_far & distance \\ \hline
how\_old & age \\ \hline
\end{tabular}
}
\caption{Question Types and Their Corresponding Key Terms}
\label{tab:keywords}
\end{table}

\begin{table}[t]
\centering
\resizebox{0.5\textwidth}{!}{
\begin{tabular}{l|c|l}
\hline
Type & Count & Question Examples \\
\hline
how & 2 & how are leaders of the two parties in congress chosen \\
what & 98 & what is the setting of the story sorry wrong number \\
which & 22 & which domain of life are humans members of \\
why & 0 & - \\
other & 127 & latest season on keeping up with the kardashians \\ \hline
total & 1667 & - \\
\hline
\end{tabular}
}
\caption{Summary of Excluded Question Types in Memory Answer and Counter Answer Generation. 
The table lists question types that were excluded from processing due to either the difficulty in identifying a unified entity type (``how'', ``what'', ``which'') or not question (``other'').}
\label{tab:exclude_quesition_types}
\end{table}


\subsection{Dataset Details}

\begin{table*}
\centering
\resizebox{1.0\textwidth}{!}{
\begin{tabular}{l|c|c|c|c|c|c|c|c|c|c|c|c}
\hline
\textbf{} & \multicolumn{6}{c|}{\textbf{popQA}} & \multicolumn{6}{c}{\textbf{NQ}} \\ \cline{2-13}
\textbf{} & \textbf{LLaMA2-7B} & \textbf{LLaMA2-70B} & \textbf{LLaMA3.2-3B} & \textbf{ChatGPT} & \textbf{GPT-4} & \textbf{Claude3.5} & \textbf{LLaMA2-7B} & \textbf{LLaMA2-70B} & \textbf{LLaMA3.2-3B} & \textbf{ChatGPT} & \textbf{GPT-4} & \textbf{Claude3.5} \\ \hline
\textbf{Initial (\# of Q)} & 1000 & 1000 & 1000 & 1000 & 1000 & 1000 & 1667 & 1667 & 1667 & 1667 & 1667 & 1667 \\ \hline
\textbf{Generate MA} & 1000 & 1000 & 1000 & 1000 & 1000 & 1000 & 1435 & 1392 & 1352 & 1532 & 1539 & 1482\\ \hline
\textbf{Generate CMA} & 1000 & 1000 &  1000 & 1000 & 1000 & 1000 & 1152 & 1101 & 1140 & 1189 & 1252  & 1232\\ \hline
\textbf{CMA filter} & 922 & 932 & 942 & 944 & 946 & 934 & 1060 & 1027 & 1123 & 1110 & 1188 &  1228\\ \hline
\rowcolor[gray]{0.9}
\textbf{Direct Evidence} & 918 & 922 & 938 & 933 & 933 & 931 & 1042 & 1009 & 1002 & 1079 & 1171 & 1173\\ \hline
\textbf{2 sentence indirect evidence} & 903 & 910 & 922 & 921 & 923 & 914 & 990 & 985 & 965 & 1038 & 1129 & 1116 \\ \hline
\textbf{3 sentence indirect evidence} & 907 & 897 & 925 & 918 & 924 & 920 &  991 & 982 & 980 & 1041 & 1125 & 1122 \\ \hline
\rowcolor[gray]{0.9}
\textbf{intersection of 2\&3 sentence evidence} & 901 & 895 & 917 & 911 & 918 & 913 & 976 & 972 & 941 & 1025 & 1108 & 1059 \\ \hline
\end{tabular}
}
\caption{The dataset scale at each step across popQA and NQ dataset. ``intersection of 2\&3 sentence evidence'' is the count for indirect evidence.}
\label{tab:evidence_count}
\end{table*}

For the popQA dataset, we use the dataset from \citet{xie2024knowledgeconflict} by randomly selecting 1,000 questions from the data intersection of the conflicts generated by LLaMA2-7B, LLaMA2-70B, LLaMA3.2-3B, ChatGPT, GPT-4 and Claude3.5. We use the MA and CMA from \citet{xie2024knowledgeconflict} and only generate direct evidence and indirect evidence using our framework. For the NQ dataset, we use the test set from \citet{longpre2021entity}, which consists of 1,667 unique questions. The MA, CMA, and evidence are all generated with our framework. The dataset scale at each step is presented in Table \ref{tab:evidence_count}. 

\subsection{Human Evaluation for Model Reliability}

To ensure the reliability of the NLI model, \citet{xie2024knowledgeconflict} randomly sample 200 generated examples and manually annotate whether the generated content entails the corresponding claim. The labels are supportive (entailment in
the NLI task) or not supportive (either neutral or contradiction in the NLI task). The accuracy is 99\%.

Following this process, we evaluate how reliable the generated CMA is. We randomly sample 200 generated examples in the NQ dataset and manually annotate whether the correct entity in MA is found and replaced with a same type alternative. The accuracy is 98\%, which means the generated CMA is reliable.

\section{Additional Studies}\label{sec:additional_studies}

\subsection{Memory Strength on Different Datasets}
\label{appendix: memory strength on datasets}

\begin{figure*}[t]
    \centering
    \includegraphics[width=\linewidth]{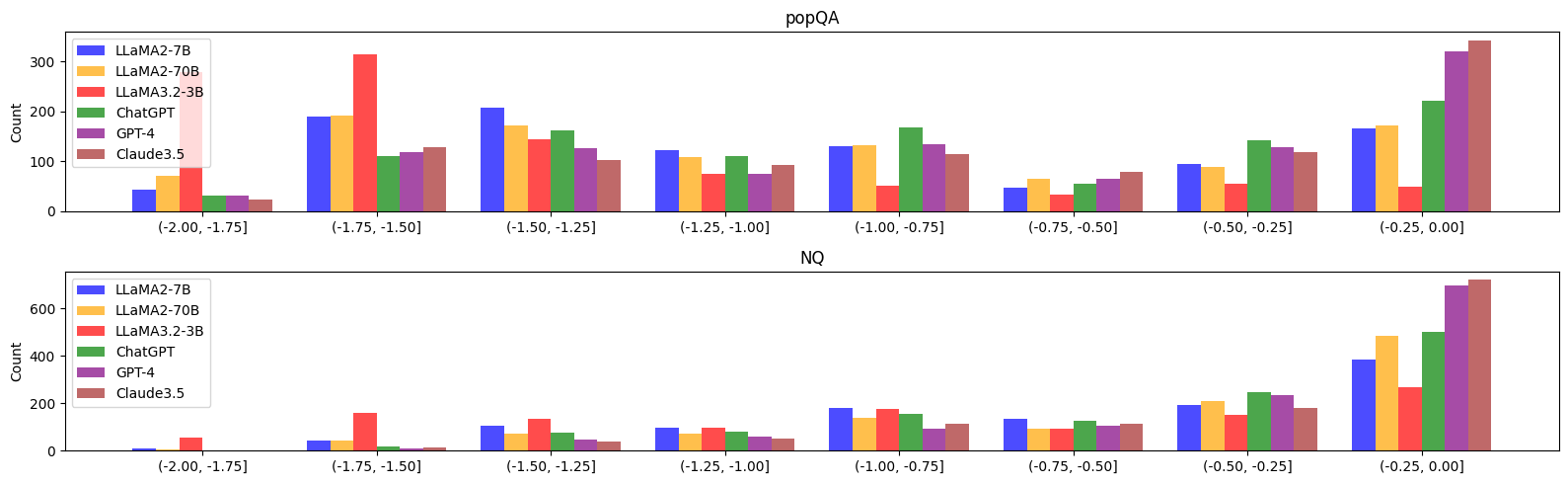}
    \caption{Memory Strength Distribution Across popQA and NQ Datasets. Each dataset is classified into 8 bins. The x-axis shows the range of memory strength for each bin. The y-axis shows the question count in each bin. The NQ dataset exhibits higher overall memory strength. Additionally, larger models (e.g., GPT-4) show stronger memory strength compared to smaller models.}
    \label{fig:memory_strength_distribution}
\end{figure*}

We illustrate the distributions of memory strength on the popQA and NQ datasets for LLaMA2-7B, LLaMA2-70B, LLaMA3.2-3B, ChatGPT, GPT-4 and Claude3.5, respectively (shown in Figure \ref{fig:memory_strength_distribution}). The results show that
\textbf{LLMs demonstrate stronger memory for the NQ dataset than the popQA dataset.} For the NQ dataset, most questions fall within the bin of (0.25, 0]. Only a few questions fall within bins of weaker memory strength. In contrast, the popQA dataset has a greater number of questions in bins with weaker memory strength. This phenomenon is consistent across all six evaluated LLMs, indicating that the LLMs have more knowledge of the NQ dataset compared to the PopQA dataset. A possible explanation is that the NQ dataset covers more commonly discussed subjects than those in the PopQA dataset. These subjects may have been encountered more frequently during the training of the LLMs, making it easier for the models to recall the information and resulting in stronger memory strength.

\subsection{Order of Options}


\begin{table*}[t]
\centering
\resizebox{\textwidth}{!}{%
\begin{tabular}{c|l|c|ccc|ccc|ccc|ccc|ccc|ccc}
\hline
    \multirow{2}{*}{Dataset} & \multirow{2}{*}{Evidence Style} & \multirow{2}{*}{S \#}  & \multicolumn{3}{c}{LLaMA2-7B} & \multicolumn{3}{c}{LLaMA2-70B} & \multicolumn{3}{c}{LLaMA3.2-3B} & \multicolumn{3}{c}{ChatGPT} & \multicolumn{3}{c}{GPT-4}  & \multicolumn{3}{c}{Claude3.5}   \\ 
    \cline{4-21}
              &  & & $R_m\downarrow$ & $R_c\uparrow$ & $R_u$ & $R_m\downarrow$ & $R_c\uparrow$ & $R_u$ & $R_m\downarrow$ & $R_c\uparrow$ & $R_u$ & $R_m\downarrow$ & $R_c\uparrow$ & $R_u$ & $R_m\downarrow$ & $R_c\uparrow$ & $R_u$ & $R_m\downarrow$ & $R_c\uparrow$ & $R_u$ \\ \hline
             & \multicolumn{20}{c}{\cellcolor[gray]{0.9}MA first} \\ \cline{2-21}
\multirow{4}{*}{popQA} & Direct & 1 & 0.44 & 99.56 & 0.0 & 1.08 & 98.7 & 0.22 & 7.69 & 85.15 & 7.16 & 3.32 & 94.75 & 1.93 & 13.29 & 84.57 & 2.14 & 0.65 & 83.31 & 16.04  \\ \cline{3-21}
& Direct + Paraphrase & 3 & 0.11 & 99.89 & 0.0 & 0.43 & 99.35 & 0.22 & 1.07 & 97.86 & 1.07 & 1.39 & 98.28 & 0.32 & 1.29 & 98.5 & 0.21 & 0.11 & 98.06 & 1.83  \\ \cline{2-21}
& \multicolumn{20}{c}{\cellcolor[gray]{0.9} CMA first} \\ \cline{2-21}
& Direct  & 1 & 59.37 & 40.41 & 0.22 & 5.75 & 93.27 & 0.98 & 18.8 & 69.12 & 12.07 & 6.22 & 91.96 & 1.82 & 21.44 & 75.24 & 3.32 & 0.43 & 76.96 & 22.61 \\ \cline{2-21}
& Direct + Paraphrase & 3 & 17.49 & 82.51 & 0.0 & 1.74 & 98.05 & 0.22 & 14.64 & 79.92 & 5.45 & 1.82 & 97.75 & 0.43 & 2.79 & 96.78 & 0.43 & 0.21 & 96.88 & 2.91  \\ \hline 
             &  \multicolumn{20}{c}{\cellcolor[gray]{0.9}MA first} \\  \cline{2-21}
\multirow{4}{*}{NQ} & Direct & 1 & 7.2 & 92.8 & 0.0 & 3.07 & 96.93 & 0.0 & 26.41 & 59.88 & 13.71 & 19.46 & 75.16 & 5.38 & 50.04 & 47.99 & 1.96 & 1.96 & 56.4 & 41.64 \\ \cline{3-21}
& Direct + Paraphrase & 3 & 3.26 & 96.74 & 0.0 & 1.19 & 98.61 & 0.2 & 9.38 & 83.27 & 7.36 & 9.55 & 86.75 & 3.71 & 11.27 & 87.28 & 1.45 & 0.26 & 93.09 & 6.65 \\ \cline{2-21} 
& \multicolumn{20}{c}{\cellcolor[gray]{0.9}CMA first} \\ \cline{2-21}
& Direct  & 1 & 22.26 & 77.73 & 0.0 & 19.13 & 80.38 & 0.5 & 41.33 & 42.44 & 16.23 & 34.48 & 61.82 & 3.71 & 49.19 & 47.99 & 2.82 & 4.78 & 39.85 & 55.38 \\ \cline{2-21}
& Direct + Parapharse & 3 & 4.8 & 95.11 & 0.1 & 8.72 & 90.39 & 0.89 & 24.19 & 65.62 & 10.18 & 18.63 & 78.96 & 2.41 & 17.76 & 80.02 & 2.22 & 1.28 & 78.84 & 19.88 \\ \hline

\end{tabular}
}
\caption{Results of LLM Receptiveness to Different Evidence Styles Across popQA and NQ Datasets.
The table presents the MA ratio ($R_m$), CMA ratio ($R_c$), and UCT ratio ($R_u$) for Direct Evidence and Direct + Paraphrase Evidence with CMA first and MA first scenarios. All the ratios are in \%.}
\label{tab:option_order}
\end{table*}

\begin{figure*}[t]
    \centering
    \includegraphics[width=0.9\linewidth]{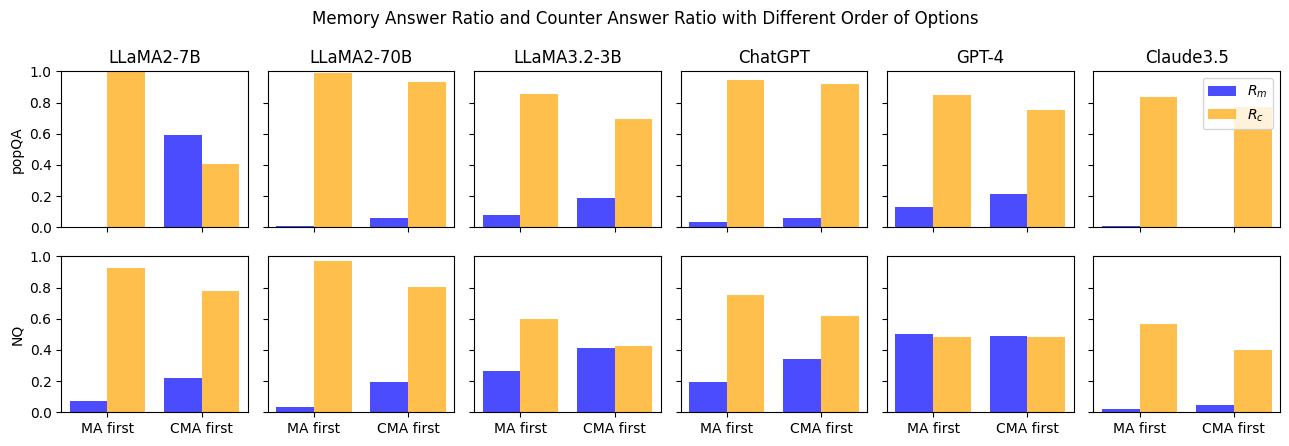}
    \caption{Impact of Option orders on Memory and Counter Ratios Across NQ and popQA Datasets. Either the memory answer ("mem first") or the counter answer ("ctr first") is introduced first to six models.}
    \label{fig:option_orders}
\end{figure*}

To test the effect of the order of options on $R_m$, we conduct an experiment with one sentence direct evidence by changing the order of options (MA option and CMA option). We define the scenario where the CMA option is presented first in the prompt as ``CMA first'', and the scenario where the MA option is presented first as ``MA first''\footnote{All previous evaluations are under ``MA first'' conditions.}. Figure \ref{fig:option_orders} shows the results.

Across all six models (LLaMA2-7B, LLaMA2-70B, LLaMA3.2-3B, ChatGPT, GPT-4 and Claude3.5), we observe a consistent trend: MA ratio ($R_m$) under ``CMA first'' is significantly higher than that under ``MA first''. 
Evaluations under ``CMA first'' demonstrate that LLMs are less context-faithful. 

To further demonstrate the effect of the order of options on $R_m$, we compare the performance of experiments with ``CMA first'' and ``MA first'' under two evidence styles: Direct Evidence with one sentence and Direct + Paraphrase with three sentences. The results are presented in Table \ref{tab:option_order}. The results show that, for different evidence styles, $R_m$ is higher in the ``CMA first'' compared to the ``MA first''. Comparing the results under the ``CMA first'', the $R_m$ of Direct + Paraphrase with three sentences is significantly lower than that of Direct Evidence with one sentence. This demonstrates that paraphrasing direct evidence is an effective method for decreasing $R_m$. Our conclusion remains unchanged.


\subsection{Entity Type}

To investigate how entity type influences context faithfulness, we categorize NQ questions into three types based on their expected answer entities: PER (person), LOC (location), and TIM (time), following the classification summarized in Table 3. Specifically:

\begin{itemize}
    \item \textbf{PER} includes questions such as ``who'', ``what\_name'', ``who\_writes'', ``who\_sings'', ``who\_wins'', and ``who\_plays''.
    \item \textbf{LOC} includes ``where'' and ``which\_country''.
    \item \textbf{TIM} includes ``when'', ``what\_year'', and ``which\_year''.

\end{itemize}

We then analyze the MA ratio ($R_m$) across different entity types using ChatGPT and LLaMA2-7B as representative models. For each category, we compute the question count, average memory strength, and MA ratio. Results are shown in Table \ref{tab:entity_type}. We observe the following patterns:

\begin{table}[ht]
\centering
\resizebox{0.5\textwidth}{!}{
\begin{tabular}{lcccc}
\toprule
\textbf{LLM} & \textbf{Entity} & \textbf{Count} & \textbf{Mem. Strength} & \textbf{MA Ratio (\%)} \\
\midrule
\multirow{3}{*}{ChatGPT} 
& PER & 664 & -0.4467 & 24.54 \\
& LOC & 107 & -0.5033 & 19.62 \\
& TIM & 217 & -0.3244 & 9.21 \\
\midrule
\multirow{3}{*}{LLaMA2-7B} 
& PER & 680 & -0.6149 & 8.52 \\
& LOC & 95 & -0.4344 & 9.47 \\
& TIM & 174 & -0.4359 & 1.15 \\
\bottomrule
\end{tabular}
}
\caption{Impact of different entity types (PER: Person, LOC: Location, TIM: Time) on memory strength and MA ratio for ChatGPT and LLaMA2-7B.}
\label{tab:entity_type}
\end{table}

\begin{table}[ht]
\centering
\resizebox{0.4\textwidth}{!}{
\begin{tabular}{lcccc}
\toprule
\textbf{LLM} & \textbf{low} & \textbf{mid-low} & \textbf{mid-high} & \textbf{high} \\
\midrule
ChatGPT & 4.76 & 9.09 & 9.09 & 10.08 \\
LLaMA2-7B & 0.00 & 2.13 & 0.00 & 2.53 \\
\bottomrule
\end{tabular}
}
\caption{MA ratios (\%) on TIM-type questions across different memory strength levels.}
\label{tab:tim_ma_by_strength}
\end{table}

\begin{itemize}
    \item For PER and LOC, which involve textual entities, higher memory strength corresponds to higher MA ratios. This is consistent with the trend reported in Section 4.2.1, where stronger memory increases the likelihood of memory-aligned responses.
    
    \item For TIM, which involves numerical entities, the MA ratio is substantially lower than for textual entities. This finding aligns with observations made in \citet{longpre2021entity}.

    \item We further divided TIM into four groups based on memory strength levels: low, mid-low, mid-high, and high. The results shown in Table \ref{tab:tim_ma_by_strength} and these results demonstrate a positive correlation between memory strength and MA ratio, which aligns with the conclusion presented in Section \ref{sec:faith_with_memory_strength}.
\end{itemize}

These results suggest that entity type, particularly the distinction between textual and numerical entities, affects context faithfulness, and that memory strength remains a key influencing factor across different entity types.

\subsection{Case Study}
Poor reasoning ability may be one of the reasons that LLMs provide MA. 
In our case study, we prompt LLMs to provide a rationale alongside the answer. We discover that some LLMs seem to exhibit reasoning errors. Below is an example that shows this phenomenon. 
\begin{quote}
\begin{spacing}{1}
\small 
\texttt{%
\textbf{Information}: Michigan State University holds the record for the highest number of Big Ten football championships, with a total of 36 titles. \\
\textbf{Question}: who has the most big ten championships in football  \\
\textbf{Options}: \\
A: Ohio State University has the most Big Ten championships in football with 36 titles. \\
B: Michigan State University has the most Big Ten championships in football with 36 titles. \\
C: Uncertain. \vspace{1mm} \\
\textbf{Answer}: \\
A: Ohio State University has the most Big Ten championships in football with 36 titles. \vspace{0.5mm} \\
According to the information provided, Michigan State University holds the record for the highest number of Big Ten football championships, with a total of 36 titles. Therefore, option A is the correct answer.
}
\end{spacing}
\end{quote}
In this case, option A represents MA, and option B represents CMA. While the model select MA as the final answer, its rationale indicates that it successfully received and processed the information from CMA. However, for some unknown reason, it still provides the MA as the final answer. This reasoning process is highly confusing. We suspect that the cause of MA can be attributed to LLMs' weakness in reasoning ability. This finding raises an interesting question about the relationship between reasoning ability and memory strength of LLMs. We leave this for future work.

\subsection{Impact of Memory Strength with Different Evidence Styles}

\begin{figure*}
    \centering
    \includegraphics[width=.95\linewidth]{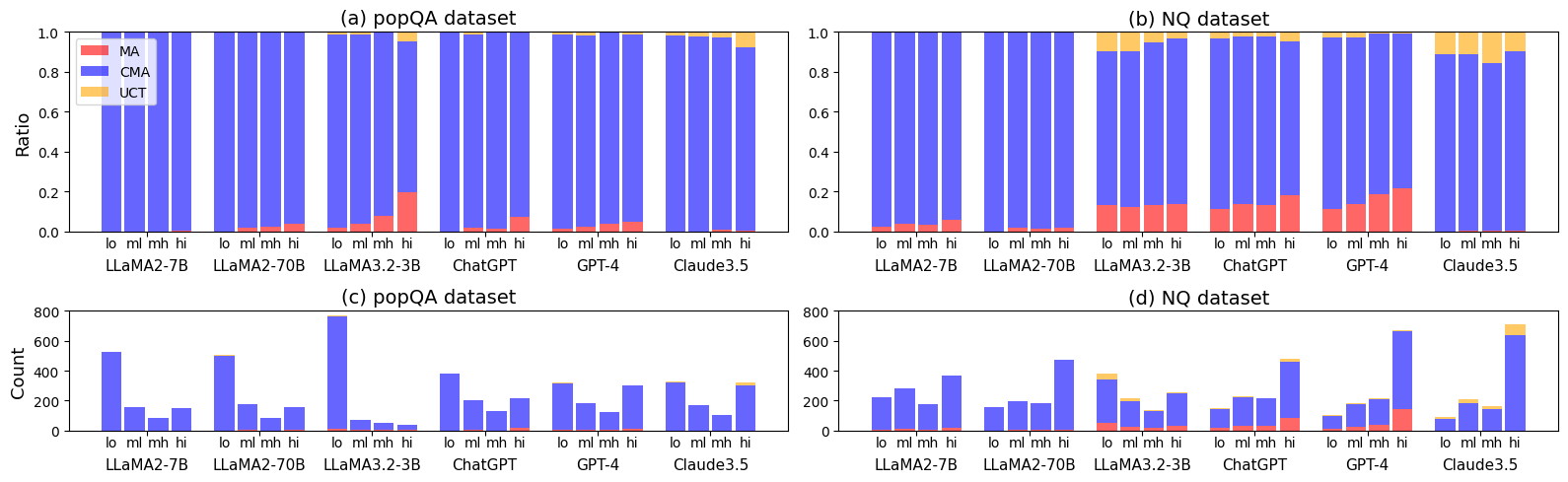}
    \caption{Relationship between Memory Strength and Different Answers' Ratios with Direct + Paraphrase Evidence with Two Sentences.}
    \label{fig:memory_strength_memory_ratio_dual_paraphrase}
\end{figure*}

\begin{figure*}
    \centering
    \includegraphics[width=.95\linewidth]{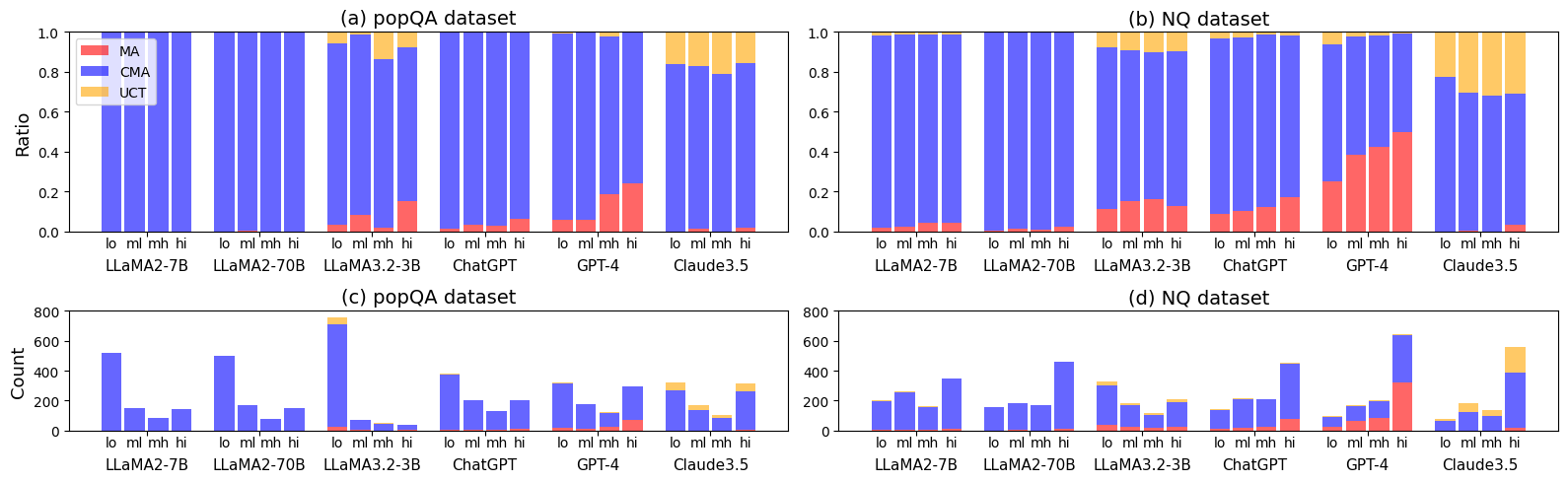}
    \caption{Relationship between Memory Strength and Different Answers' Ratios with Indirect Evidence with Two Sentences.}
    \label{fig:memory_strength_memory_ratio_two_sentence}
\end{figure*}

To demonstrate the relationship between context-faithfulness and memory strength with other evidence styles, we categorize the questions in each dataset into four groups according to the memory strength intervals [-2, -1], (-1, -0.5], (-0.5, -0.25] and [-0.25, 0], The evidence styles are direct + paraphrase evidence with two sentences and indirect evidence with two sentences. Figures \ref{fig:memory_strength_memory_ratio_dual_paraphrase},\ref{fig:memory_strength_memory_ratio_two_sentence} show the result. The figures show that there is a clear positive correlation between memory strength and MA ratio for both evidence styles, which implies that this positive correlation between memory strength and MA ratio is general.

\begin{figure*}
    \centering
    \includegraphics[width=.95\linewidth]{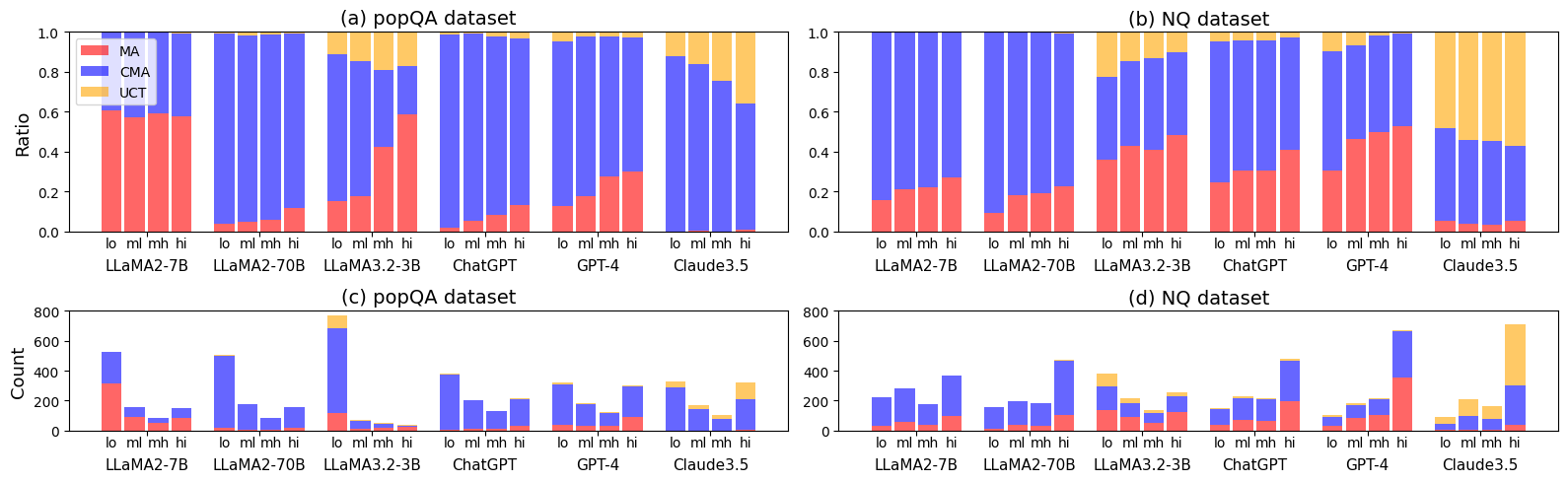}
    \caption{Relationship between Memory Strength and Different Answers' Ratios with Direct Evidence. The option order is Counter First.}
    \label{fig:memory_strength_memory_ratio_dircet_counter}
\end{figure*}

To demonstrate the relationship between context-faithfulness and memory strength with ``CMA first'' scenario, we show MA, CMA, and UCT ratios with direct evidence with one sentence under ``CMA first'' scenario in Figure \ref{fig:memory_strength_memory_ratio_dircet_counter}. The positive correlation between memory strength and MA ratio stays unchanged. 

\section{Prompts}\label{sec:prompts}
In Table \ref{tab:prompts}, we present a detailed list of all the prompts used throughout this study.

\begin{table*}
    \centering
    \resizebox{0.98\textwidth}{!}{
    \begin{tabular}{p{1.5cm}|p{1cm}|p{2.5cm}|p{12cm}}
    \hline
\textbf{Step} & \textbf{index} & \textbf{Prompt Name} & \textbf{Prompts} \\ \hline
 \multirow{15}{1.5cm}{Step \text{1:} Memory Strength} & \multirow{3}{1cm}{\centering1} & \multirow{3}{2.5cm}{Question paraphrase prompt} & Generate 7 meaningful paraphrases for the following question: [Question]. Read the question carefully.  \\ 
 & & & Paraphrases: \\ \cline{2-4}
 & \multirow{6}{1cm}{\centering2} & \multirow{6}{2.5cm}{Question equivalent check prompt} & Determine whether the paraphrased question describes the same thing as the original question, and give "Contradicted" if they are not the same, otherwise give "Same" as the result. \\ 
& & & Q1: [Paraphrased Q1] \\
& & & Q2: [Paraphrased Q2] \\
& & & Keep the answer short and concise. \\ \cline{2-4}
& \multirow{10}{1cm}{\centering3} & \multirow{10}{2.5cm}{Answer consistency check prompt} & Determine whether the answer `A1' is `Contradicted' or `Same' with the answer `A2' for the question `Q'. You need to check whether the two answers exactly have the same answer to the question. The answer could be person, name, place, time, number, genre, occupation, sport, entity, digit, or arithmetical results. If the two answers are the same, give ``Same'', otherwise give ``Contradicted'' as the result. \\
& & & Q: [question]  \\
& & & A1: [LLM answer 1] \\
& & & A2: [LLM answer 2] \\
& & & Keep the answer short and concise. \\ \hline
\multirow{12}{1.5cm}{Step \text{2:} MA and CMA} & \multirow{6}{1cm}{\centering4} & \multirow{6}{2.5cm}{Close book QA prompt} & Answer the question with one sentence with object and subject only. Give a statement that is most likely to be true directly. \\
& & & \\
& & & Question: \\
& & & [Question] \\
& & & Answer: \\ \cline{2-4}
& \multirow{5}{1cm}{\centering5} & \multirow{5}{2.5cm}{Change MA to CMA prompt} & Context: \\
& & & [CMA] \\
& & & Change the [entity type] part of the context. When multiple parts need to be changed, only choose one part to change. \\
& & & Answer: \\ \hline
\multirow{12}{1.5cm}{Step \text{3:} Evidence} & \multirow{4}{1cm}{\centering6} & \multirow{4}{2.5cm}{Direct evidence prompt} & Please paraphrase the following sentence by changing the terms, order, and phrases, but keep the meaning the same. \\
& & & \\
& & & Sentence: [CMA] \\ \cline{2-4}
& \multirow{7}{1cm}{\centering7} & \multirow{7}{2.5cm}{Indirect evidence prompt} & Given a claim, please write a short piece of detailed evidence to support it. Please ignore the correctness of the claim. You can make up fake content and supporting evidence but it should be as realistic as possible. \\
& & & Claim:  \\
& & & [counter memory answer] \\
& & & Evidence: \\
& & & Give the answer in [2 or 3] sentences directly. \\ \hline
\multirow{8}{1.5cm}{Step \text{4:} Evaluation} & \multirow{8}{1cm}{\centering8} & \multirow{8}{2.5cm}{Evaluate with evidence prompt} & According to the given information, choose the best choice from
the following options. \\
& & & Information: [evidence] \\ 
& & & Question: [question] \\
& & & Option: \\ 
& & & A: [option 1] \\
& & & B: [option 2] \\
& & & ... \\
& & & Answer: \\ \hline
    \end{tabular}
    }
    \caption{Prompts for LLMs in this paper. ``[PLACEHOLDER]'' is the corresponding input. }
    \label{tab:prompts}
\end{table*}

\end{document}